\documentclass[conference]{IEEEtran}
\usepackage[pscoord]{eso-pic}
\IEEEoverridecommandlockouts

\usepackage{cite}
\usepackage[font=footnotesize,skip=3pt]{caption}
\usepackage{amsmath,amssymb,amsfonts}
\usepackage{dsfont}
\usepackage{algorithmic}
\usepackage{graphicx}
\usepackage{subcaption}
\usepackage{booktabs}
\usepackage{array}
\usepackage{multirow}
\usepackage{textcomp}
\usepackage{xcolor}
\usepackage{comment}
\usepackage{siunitx}
 \usepackage[font=footnotesize]{caption}
\usepackage[colorlinks=true, allcolors=blue]{hyperref}
\usepackage{orcidlink}

\def\BibTeX{{\rm B\kern-.05em{\sc i\kern-.025em b}\kern-.08em
    T\kern-.1667em\lower.7ex\hbox{E}\kern-.125emX}}
\begin{document}
 
\title{Explanation-Based Runtime Verification for Trustworthy ML-driven Optical Networks\thanks{This work has been supported by the EUREKA CELTIC-NEXT SUSTAINET-Advance project, funded by the Swiss Innovation Agency Innosuisse No. 119.588 INT-ICT, and by Vinnova (Sweden's Innovation Agency) No. 2025-02987. Corresponding author: omran.ayoub@supsi.ch}}
\author{
\IEEEauthorblockN{
Omran Ayoub~\orcidlink{0000-0002-3884-3594}\IEEEauthorrefmark{1}
Carlos Natalino~\orcidlink{0000-0001-7501-5547}\IEEEauthorrefmark{2},
Ali Al Housseini~\orcidlink{0009-0003-6682-474X}\IEEEauthorrefmark{1},
Felix Foschum\IEEEauthorrefmark{3}, \\
Philipp Morger\IEEEauthorrefmark{3}, 
Tiziano Leidi~\orcidlink{0000-0002-6335-7977}\IEEEauthorrefmark{1},
David Hock~\orcidlink{0009-0002-5142-2687}\IEEEauthorrefmark{4},
Paolo Monti~\orcidlink{0000-0002-5636-9910}\IEEEauthorrefmark{2}
}
\IEEEauthorblockA{\IEEEauthorrefmark{1}
University of Applied Sciences and Arts of Southern Switzerland, Lugano, Switzerland}
\IEEEauthorblockA{\IEEEauthorrefmark{2}
Department of Electrical Engineering, Chalmers University of Technology, Gothenburg, Sweden}
\IEEEauthorblockA{\IEEEauthorrefmark{3} SKOOR GmbH, Switzerland, \IEEEauthorrefmark{4}
Infosim GmbH \& Co. KG, Germany}
}

\maketitle

\begin{abstract}
Machine learning (ML) models are increasingly integrated into optical network automation frameworks to support tasks such as failure management, performance monitoring and resource allocation. In these environments, ML-driven predictions may be directly coupled with control-plane actions where incorrect decisions can immediately impact service quality, resource efficiency, and network stability. As automation levels increase, ensuring the reliability of individual decisions at deployment time becomes a critical requirement. 
Explainable artificial intelligence (XAI) techniques have emerged to improve transparency by highlighting the factors influencing ML predictions. In addition to identifying influential features, they provide insights into the underlying reasoning process of the model, revealing how different input variables contribute to the final outcome and how feature interactions shape the decision boundary. In this work, we introduce explanation-based runtime verification, an approach that exploits model explanations to assess the soundness of individual ML decisions before they are executed in the network control loop. The proposed approach evaluates explanation coherence and physics grounding consistency at runtime, enabling the system to defer or reject decisions flagged as uncertain. We demonstrate the effectiveness of our approach on a representative use case of lightpath quality of transmission classification. Experimental results show that explanation-based verification can intercept a significant fraction of erroneous decisions while preserving high automation rate.
\end{abstract}

%

\section{Introduction}
Machine learning (ML) models are increasingly deployed in optical networks to support operational decisions such as configuration selection, performance estimation, anomaly detection, and resource allocation \cite{castoldi2025shaping}. As these models become embedded in control and orchestration systems, their outputs are no longer merely advisory but they directly influence network actions. In such environments, incorrect ML decisions can lead to service degradation, inefficient resource utilization, or instability in automated workflows \cite{natalino2024ai}.

Ensuring high aggregate predictive accuracy during offline validation is not sufficient to guarantee safe deployment as such models may still produce isolated erroneous decisions under specific network conditions. From an operator's perspective, the critical question is therefore not only \emph{how accurate is the model overall?}, but rather \emph{can a specific decision be trusted at runtime before it is enacted?} Avoiding wrong automated decisions, particularly false-positive ones that may trigger unsafe configurations, is therefore essential in operational optical networks.

Explainable artificial intelligence (XAI) techniques have been introduced to improve transparency by revealing which input features influence a model's prediction, with the ultimate goal of enabling model behavior verification and enhancing trust \cite{dwivedi2023explainable}. While such explanations enhance interpretability and facilitate offline analysis, they do not inherently provide mechanisms to assess the reliability of individual decisions at inference time. An explanation can describe a decision, but it does not verify whether the reasoning is coherent, plausible, or sufficiently supported to justify operational enforcement.

In this work, we introduce \emph{explanation-based runtime verification} for ML-driven decision systems in optical networks. Runtime verification, in this context, refers to the continuous assessment of ML outputs against predefined acceptability criteria, enabling the system to selectively accept decisions that are well-supported and defer those that exhibit contradictory or insufficient explanatory evidence. In other words, we use explanations as runtime evidence to verify the decision of the underlying ML model.

\newcommand{\placetextbox}[3]{
  \setbox0=\hbox{#3}
  \AddToShipoutPictureFG*{
    \put(\LenToUnit{#1\paperwidth},\LenToUnit{#2\paperheight})%
    {\vtop{{\null}\makebox[0pt][c]{#3}}}}
  }
\placetextbox{.2}{0.055}{978-3-903176-78-2 \textcopyright\ 2026 IFIP}

We first define an uncertainty criterion to identify potentially unreliable or borderline predictions (e.g., low confidence or small margin between classes). For each such flagged instance, we extract feature attributions using an XAI technique, namely, Shapley Additive Explanations (SHAP) \cite{lundberg2017unified}, and analyze the direction and magnitude of feature influence at the individual decision level. These attributions are then evaluated against predefined acceptability rules that assess both decision–explanation coherence (i.e., whether the dominant features genuinely support the predicted class) and consistency with optical-layer physics (i.e., whether the influential features align with known physical principles and operational knowledge). If the dominant explanatory factors consistently support the predicted outcome and correspond to meaningful network drivers, the decision, despite being uncertain, is accepted. Otherwise, it is deferred to conservative handling procedures or fallback mechanisms, such as human supervision. The objective is therefore to selectively identify and suppress unreliable uncertain decisions, rather than indiscriminately discarding all low-confidence predictions, so as to maintain a high level of automation while significantly mitigating the risk of erroneous decisions.

We hypothesize that predictions exhibiting incoherent, unstable, or physically inconsistent explanatory patterns are inherently more prone to being unreliable or spurious. By validating the reasoning structure underlying each prediction, the system can identify decisions whose internal logic conflicts with domain principles or operational knowledge, thereby reducing the risk of incorrect control actions. The overall goal is to strengthen the trustworthiness and operational safety of ML-driven optical network automation through explanation-aware consistency checks embedded directly in the deployment pipeline, without altering the underlying predictive model or imposing significant computational overhead.

We demonstrate the effectiveness of our approach on a representative use case of binary lightpath Quality of Transmission (QoT) classification. Experimental results on two open datasets show that the proposed approach supports the selective rejection of high-risk approvals, lowering the false positive rate while sustaining a high level of automation.

\section{Preliminaries: Feature Attribution}
This section introduces the fundamental concepts of XAI and feature attribution, as the latter constitutes a core building block of our approach. 

XAI encompasses methods and frameworks designed to make the decision-making processes of ML and AI models transparent and understandable to humans. Its primary objective is to \lq \lq open'' black-box models by generating interpretable explanations that reveal how input features influence predictions. Beyond transparency, XAI allows practitioners to uncover relationships learned by the model, identify influential input features and verify alignment with domain knowledge. It further supports model debugging and human oversight in critical scenarios.

The most widely adopted XAI approaches are post-hoc explanation methods. These techniques are applied after model training, without modifying the internal structure or predictive behavior of the model, and aim to interpret how a given decision has been reached. Depending on their scope, post-hoc explanations can provide either global or local insights. Global explanations characterize the overall behavior of the model across the dataset, for example by revealing the influence of each feature and verifying consistency with known trends. Local explanations, in contrast, focus on a single prediction and identify the specific features that contributed most to that individual outcome.

\begin{figure}
    \centering
    \includegraphics[width=0.99\linewidth]{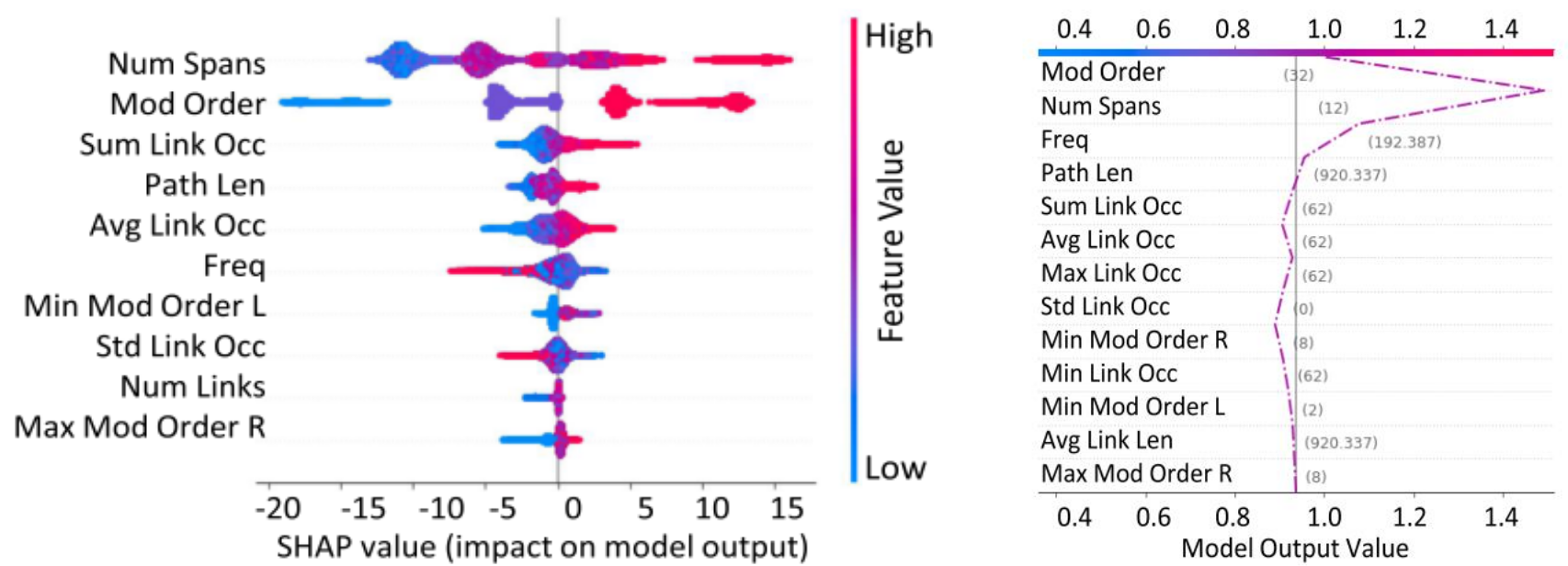}
    \caption{\footnotesize{Example of a global explanation (left) and a local explanation (right) generated using SHAP for a lightpath QoT classification model. The global explanation visualizes the overall feature impact (SHAP values) across the dataset for the unacceptable QoT class, characterizing the general behavior of the model. For example, it indicates that a higher number of spans, higher-order modulation formats, increased link occupation, and longer path length contribute to degraded lightpath QoT, in agreement with domain knowledge. The local explanation, in contrast, presents the feature attributions for a single prediction, quantifying how each input feature contributes to that specific decision. Further details on how to interpret these plots are provided in \cite{ayoub2022towards}.}}
    \label{fig:globallocal}
    \vspace{-0.5cm}
\end{figure}

In this work, we rely on local explanations and specifically on feature attribution methods, i.e., feature-based explanations that quantify the influence of each input variable on the model output. 
Figure \ref{fig:globallocal} illustrates an example of a global explanation (left) and a local explanation (right)\footnote{\cite{ayoub2022towards} provides a detailed interpretation of these plots in the context of lightpath QoT estimation}. The local explanation plot quantifies the contribution of individual features to a specific prediction, distinguishing between features that support the model’s decision (positive contribution) and those that counteract it (negative contribution). The magnitude of each contribution reflects the strength of its influence on the final output, thereby enabling a fine-grained understanding of the reasoning behind a single automated decision. In our work, we employ feature attribution to measure feature influence at the level of individual decisions. Methods such as SHAP \cite{lundberg2017unified} provide a principled way to decompose a prediction into additive feature contributions, supporting both local and aggregated analyses. This feature-level quantification of influence constitutes the foundation of our runtime verification framework, as it enables systematic assessment of whether the reasoning reflected in the attribution scores is operationally and physically consistent before a decision is enacted.

\section{Related Work}
XAI has been increasingly investigated in the context of optical networks, mainly to enhance transparency, extract operational insights, and understand the influence of input features on ML model decisions. However, existing works primarily exploit explainability for post-hoc interpretation and knowledge distillation, rather than for runtime verification or operational validation of ML-driven decisions.

Several studies focused on XAI for lightpath QoT estimation. In~\cite{ayoub2022towards}, XAI techniques were applied to a supervised QoT classification model to extract insights into feature relevance and analyze misclassifications. Similarly,~\cite{ayoub2022application,ayoub2022quantifying} leveraged SHAP-based explanations to quantify the influence of input features and identify combinations of feature values driving QoT predictions. In~\cite{fawaz2024reducing}, XAI was used to identify the most impactful features, enabling feature set reduction to lower telemetry load and computational complexity while preserving predictive performance. Along the same line,~\cite{aladin2025automated} integrated XAI-driven feature selection with transfer learning techniques to enhance model adaptability and reliability across heterogeneous datasets. Beyond technical performance,~\cite{kumar2025building} investigated interpretable AI models for optical network management tasks and assessed the impact of explainability on operator trust and fault localization.

Explainability has also been exploited for fault management. In~\cite{karandin2022if}, XAI was applied to interpret ML-based fault localization decisions using optical signal measurements. Similarly,~\cite{jenila2025performance} incorporated SHAP and Local Interpretable Model Agnostic Explanation frameworks to enhance transparency and operator trust in ML-based real-time fault detection using Optical Time Domain Reflectometer data. Moreover, in the context of traffic prediction and dynamic network optimization,~\cite{goscien2024traffic} used XAI to identify which transmission and network state parameters most influence bandwidth blocking probability in elastic optical networks, while~\cite{knapinska2024explainable} extracted insights into traffic prediction models by analyzing feature relevance across traffic types and aggregation levels. More recently, explainability has been explored in reinforcement learning (RL)-based control frameworks. In~\cite{asdikian2025verifying}, Shapley-based explanations were integrated with a deep RL agent for network slice admission control to analyze and validate rejection decisions prior to deployment. Similarly,~\cite{mehyeddine2025beyond} enhanced Shapley-based explainability for RL-driven routing and spectrum assignment to interpret proactive lightpath rejections.

Apart from explainability, several works have focused on enhancing the reliability of ML models for optical networks through uncertainty quantification (UQ) and probabilistic modeling. These approaches aim to provide calibrated confidence intervals, prediction margins, or distributional estimates to support risk-aware decision making. For instance, calibrated regression and probabilistic QoT modeling techniques were proposed in~\cite{di2022calibrated,di2023uncertainty}, while deep quantile regression and Monte Carlo dropout were leveraged in~\cite{maryam2022learning,maryam2022representing} to explicitly represent prediction uncertainty and reduce overly conservative design margins. More recently, conformal prediction frameworks have been introduced to provide statistically guaranteed QoT estimates under domain shift~\cite{rezaeipolicy}.

While these approaches quantify predictive uncertainty at the output level, they do not explicitly validate the internal reasoning structure of individual decisions. In contrast, our work leverages model explanations as runtime evidence to assess decision–explanation coherence and physics consistency. Importantly, explanation-based verification is orthogonal to uncertainty-aware modeling and can be naturally combined with UQ or conformal prediction techniques to further enhance decision reliability.

Finally, we note that a few works have explored the use of explanations for verification-oriented purposes during operation. In~\cite{barnard2022robust}, SHAP explanations were used as secondary signals to distinguish between previously seen and unseen network attacks, effectively leveraging feature attributions for robustness enhancement. Similarly,~\cite{soundrarajan2026ai} proposed lightweight interpretable verifiers to perform near-real-time consistency checks of DRL agents in Open RAN environments. While these approaches use interpretability as an auxiliary verification mechanism, they do not explicitly incorporate domain knowledge or physics-based constraints into the validation process. Our work follows a similar direction but incorporates operational and physical consistency considerations into explanation-based verification for optical networks.

\begin{figure}
    \centering
    \includegraphics[width=0.9\linewidth]{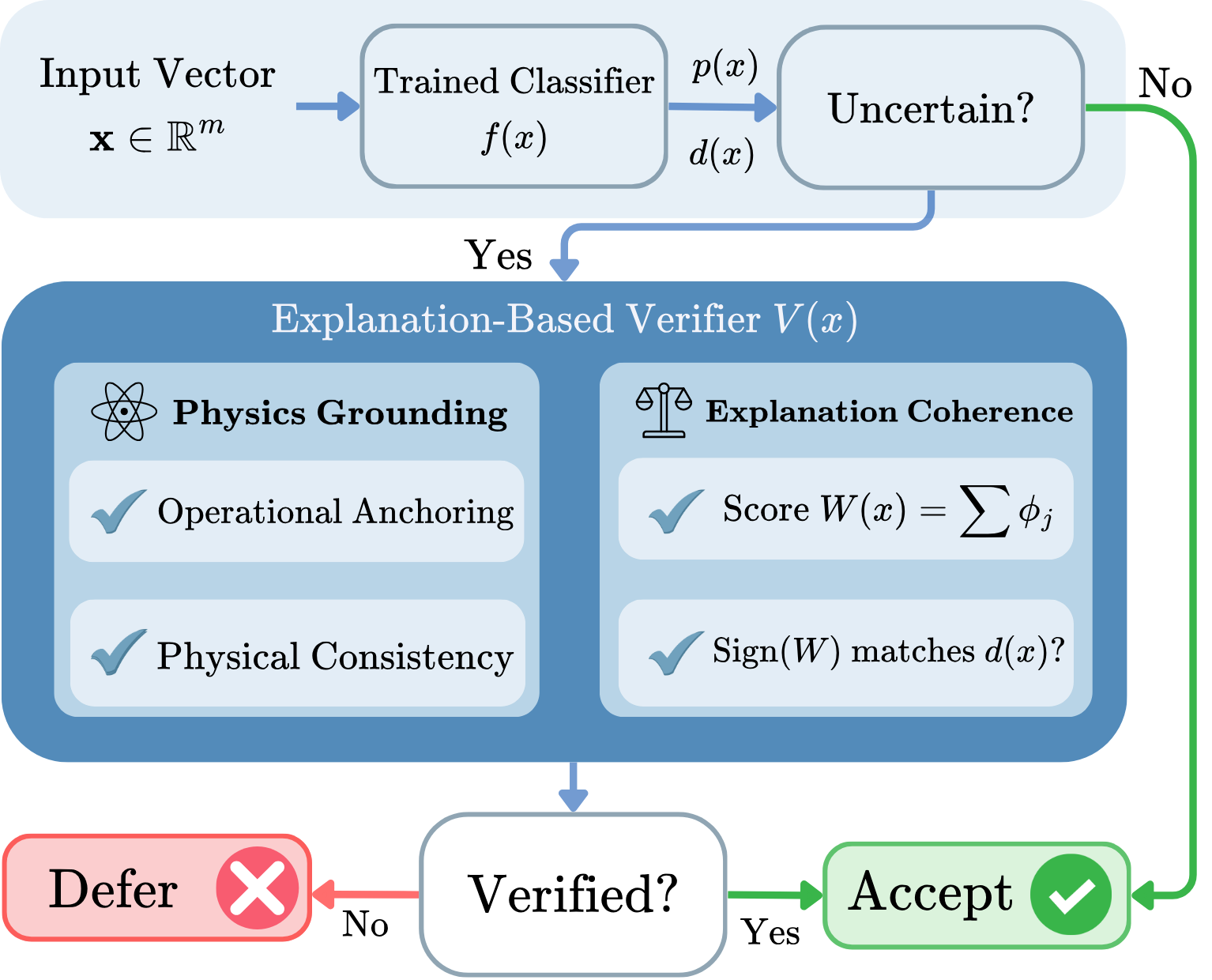}
  \caption{Schematic representation of the proposed approach.}
    \label{fig:method}
    \vspace{-0.5cm}
\end{figure}

\section{Explanation-based Runtime Verification}

The objective of our work is to introduce a runtime mechanism that evaluates, on a per-decision basis, i.e., at inference time, whether a given prediction should be trusted for execution, focusing specifically on predictions flagged as uncertain. We formalize this mechanism as a runtime verifier
\begin{equation}
V(\mathbf{x}) \in \{\texttt{ACCEPT}, \texttt{DEFER}\},
\end{equation}
which operates immediately after inference and prior to execution.

The verifier relies exclusively on information available at runtime and derived from model's explanations. The focus is on identifying weakly supported decisions before they propagate into automated network control. For a number of identified uncertain decisions, the verifier will do specific checks and then decide whether to accept the decision or to defer it. The goal is to preserve high automation rate (i.e., reduce human intervention) while reducing the error rate among enforced decisions. 

As an illustrative case study, we consider binary lightpath QoT classification, where $d(\mathbf{x})=1$ denotes a predicted feasible lightpath.

Figure \ref{fig:method} provides an overview of the proposed approach, referred to as \emph{XAI-based}, and its main components. Consider a trained binary classifier 
$f: \mathbb{R}^m \rightarrow [0,1]$ 
deployed within an optical network control system. 
Given an input vector $\mathbf{x} \in \mathbb{R}^m$, the model outputs a probability
$p(\mathbf{x})$ 
and a decision
\begin{equation}
    d(\mathbf{x}) = \mathds{1}\!\left[p(\mathbf{x}) \ge \theta\right]
    \label{eq:d}
\end{equation}
where $\theta$ is a predefined threshold. 

To identify predictions requiring additional scrutiny, we compute predictive entropy from the classifier’s output probabilities\footnote{Entropy is higher when the model assigns similar probabilities to both classes, indicating uncertainty}. Samples are ranked by entropy, and the top $q\%$ most uncertain predictions are flagged and forwarded to the runtime verifier.

For uncertain decisions, we compute a feature attribution vector
\begin{equation}
\boldsymbol{\phi}(\mathbf{x}) = [\phi_1(\mathbf{x}), \ldots, \phi_m(\mathbf{x})],
\end{equation}
where $\phi_j(\mathbf{x})$ quantifies the contribution of feature $j$ toward the positive class\footnote{We obtain attributions using SHAP but any XAI technique that provides local explanations can be used.}.
However, it is important to note that the framework can be applied to a wide range of ML tasks such as binary classification and multi-class classification.

\subsection{Verification Checks} 
We consider two verification checks.

\textbf{Physics Grounding:}
The first verification aims to ensure that decisions are supported by operationally meaningful and physically consistent evidence. We therefore introduce a \emph{physics grounding} check.

Let $\mathcal{P}$ denote the set of core operational features known to influence ML model's decision, such as path length, number of spans, modulation order, and which correspond to physically interpretable parameters directly related to signal degradation mechanisms and feasibility conditions \cite{ayoub2022towards}.

At runtime, we restrict attention to the dominant explanatory factors $\mathcal{T}_K(\mathbf{x})$, defined as the top-$K$ features ranked by absolute attribution magnitude. The grounding check enforces two complementary requirements:

\begin{itemize}
    \item \textbf{Operational anchoring:} At least one dominant explanatory feature must belong to $\mathcal{P}$. This aims to ensure that the decision is supported by core impairment-related features rather than weakly-correlated attributes.
    
    \item \textbf{Physics consistency:} For a subset of features that show a monotonic impact on model's decisions during training, the direction of their contribution at inference must align with basic physical intuition. This property was verified by inspecting the global explanations of the model (e.g., feature importance rankings and aggregated SHAP value distributions), as reported in~\cite{ayoub2022towards}.
\end{itemize}
 
Concretely, a decision is deferred if a dominant impairment-related feature appears among the top-$K$ attributions but its contribution contradicts expected monotonic behavior. For example, if a high modulation format or an unusually long path length provides strong positive attribution toward the feasible class, the decision is flagged. The intuition is that enforcing monotonic consistency constraint can reduces the likelihood that the model relies on spurious correlations or unstable statistical artifacts that contradict general domain knowledge.

\textbf{Coherence:}
The second verifiability check relates to the coherence between the model's decision and the dominant part of the explanation. It aims to enforce logical consistency between the model's decision and its own dominant explanatory evidence. 

Let $\mathcal{T}_K(\mathbf{x})$ denote the indices of the top-$K$ features ranked by absolute attribution magnitude $|\phi_j(\mathbf{x})|$. 
We define a witness score
\begin{equation}
W(\mathbf{x}) = \sum_{j \in \mathcal{T}_K(\mathbf{x})} \phi_j(\mathbf{x}).
\end{equation}

Also here, $K$ is a design parameter that determines how many dominant explanatory features are considered when assessing the decision. 
The attribution value $\phi_j(\mathbf{x})$ represents the contribution of feature $j$ to the prediction of the positive class. 
Positive values indicate that the feature pushes the decision toward class $1$, while negative values indicate that it pushes the decision toward class $0$.

Because $\phi_j(\mathbf{x})$ quantifies directional support toward the positive class, the sign of $W(\mathbf{x})$ captures the \emph{net support} provided by the dominant explanatory factors. In other words, $W(\mathbf{x})$ summarizes whether the strongest drivers (strongest features) of the prediction collectively argue in favor of feasibility or infeasibility (and consequently, in favor of the decision or not).

The coherence condition is defined as:
\begin{equation}
\begin{cases}
W(\mathbf{x}) > 0, & \text{if } d(\mathbf{x}) = 1, \\
W(\mathbf{x}) < 0, & \text{if } d(\mathbf{x}) = 0.
\end{cases}
\end{equation}

If the classifier predicts feasibility ($d(\mathbf{x})=1$), then the strongest contributing features should collectively provide positive support toward feasibility. Conversely, if the classifier predicts infeasibility ($d(\mathbf{x})=0$), the dominant features should collectively provide negative support. If the explanation contradicts the decision, i.e., the most influential features argue in the opposite direction of the predicted class, this condition is violated and hence flagged by the verifier. For example, in the lightpath QoT estimation use case, if the classifier predicts that a configuration is feasible ($d(\mathbf{x})=1$), the dominant explanatory features should indicate conditions consistent with feasibility. On the contrary, it may happen that the top contributors show strongly negative attributions, factors typically associated with signal degradation, while the final decision is feasible. Here, the strongest evidence argues against the predicted outcome, indicating incoherent reasoning. Similarly, if a configuration is predicted infeasible ($d(\mathbf{x})=0$), but the dominant features collectively provide positive contributions, the explanation does not support the decision. The intuition is that such inconsistency suggests that the decision may rely on weaker secondary effects, fragile feature interactions, or borderline threshold behavior.

\subsection{Decision Policy}
The runtime verifier combines both checks to determine whether a model decision should be enacted:
\begin{equation}
V(\mathbf{x}) =
\begin{cases}
\texttt{ACCEPT}, & \text{if both conditions hold}, \\
\texttt{DEFER}, & \text{otherwise}.
\end{cases}
\end{equation}

If the constraints are satisfied, the decision is accepted; otherwise, it is labeled as \texttt{DEFER}. Deferred decisions are not directly executed but may trigger fallback strategies, conservative configurations, additional validation, or operator review. 

\section{Experimental Evaluation}
\subsection{Experimental Setup}
To evaluate the proposed approach, we conduct experiments on two lightpath QoT estimation datasets \cite{bergk2021ml}. We employ an Extreme Gradient Boosting (XGBoost) classifier as the underlying ML model. Model hyperparameters are selected using standard validation procedures on the training folds and remain fixed across experiments.

For each dataset, we randomly sample $N=30{,}000$ instances and perform stratified repeated holdout evaluation using five independent train--test splits. In each split, 70\% of the data are used for training and 30\% for testing, while preserving the original class distribution. All reported results correspond to averages across the five splits.

We compute feature attributions using SHAP. For \emph{coherence}, we consider the top-$K$ most influential features ranked by absolute SHAP value, with $k=4$. For \emph{Physics Grounding}, we focus on three domain-relevant features: \emph{Path Length}, \emph{Number of Spans}, and \emph{Modulation Order}. Physics-related thresholds are derived from the 80th percentile of the corresponding feature distributions computed on the training folds. In these cases, if the sign of the SHAP contributions contradict expected physical behavior (e.g., high path length contributing positively to a feasible prediction beyond acceptable limits), the decision is rejected.

We compare XAI-based to two other approaches:
\begin{itemize}
    \item \textbf{Baseline:} the XGBoost classifier without any rejection mechanism.
    \item \textbf{Entropy Rejection:} a selective rejection baseline that discards the top-$q$ fraction of test samples with highest predictive entropy. This approach serves as a purely uncertainty-driven rejection strategy.
\end{itemize}

Selective rejection is controlled via an inspection budget parameter $q \in \{0.005, 0.01, 0.02, 0.05\}$, corresponding to inspection rates of 0.5\%, 1\%, 2\%, and 5\% of the test set.

We report both operational and predictive performance metrics computed on the accepted subset of samples (i.e., those not rejected by the verifier), following a selective classification setting. All results are reported as percentage improvements relative to the baseline classifier, where positive values indicate improvement (for FPR and FNR, improvement corresponds to a reduction). This evaluation protocol enables analysis of the risk--automation trade-off under different rejection strategies.

\begin{table}[t]
\centering
\scriptsize
\setlength{\tabcolsep}{4pt}
\caption{Percentage improvement over the baseline classifier (positive values indicate improvement) over Dataset 1.}
\label{tab:improvement_results}
\begin{tabular}{c c r r r r r r}
\toprule
Method & $q$ 
& $\Delta$Auto
& $\Delta$Acc
& $\Delta$TPR
& $\Delta$TNR 
& $\Delta$FPR 
& $\Delta$FNR  \\
\midrule
Baseline & -- 
& 0 & 0 & 0 & 0 & 0 & 0 \\
\midrule

\multirow{4}{*}{\rotatebox{90}{Entropy}}
& 0.005 & -0.51 & 0.23 & 0.13 & 1.61 & 44.32 & 55.35 \\
& 0.010 & -1.01 & 0.31 & 0.18 & 2.25 & 61.85 & 74.17 \\
& 0.020 & -2.06 & 0.41 & 0.23 & 2.87 & 79.07 & 97.00 \\
& 0.050 & -6.50 & 0.42 & 0.24 & 0.25 & 86.90  & 100.00 \\
\midrule

\multirow{4}{*}{\rotatebox{90}{XAI-based}}
& 0.005 & -0.20 & 0.10 & -0.00 & 1.67 & 46.08 & -0.12 \\
& 0.010 & -0.33 & 0.15 & -0.00 & 2.36 & 65.11 & -0.20 \\
& 0.020 & -0.55 & 0.19 & -0.00 & 3.03 & 83.49 & -0.37 \\
& 0.050 & -0.86 & 0.20 & -0.00 & 3.25 & 89.67 & -0.74 \\
\bottomrule
\end{tabular}
\end{table}

\begin{table}[t]
\centering
\scriptsize
\setlength{\tabcolsep}{4pt}
\caption{Percentage improvement over the baseline classifier (positive values indicate improvement) over Dataset 2.}
\label{tab:improvement_results_exp2}
\begin{tabular}{c c r r r r r r}
\toprule
Method & $q$ 
& $\Delta$Auto
& $\Delta$Acc 
& $\Delta$TPR
& $\Delta$TNR
& $\Delta$FPR
& $\Delta$FNR \\
\midrule
Baseline & -- 
& 0 & 0 & 0 & 0 & 0 & 0 \\
\midrule

\multirow{4}{*}{\rotatebox{90}{Entropy}}
& 0.005 & -0.54 & 0.27 & 0.22 & 0.38 & 23.68 & 27.41 \\
& 0.010 & -1.06 & 0.45 & 0.40 & 0.59 & 36.68 & 49.56 \\
& 0.020 & -2.04 & 0.68 & 0.65 & 0.72 & 45.12 & 81.37 \\
& 0.050 & -5.06 & 0.91 & 0.77 & 1.26 & 78.92 & 95.59 \\
\midrule

\multirow{4}{*}{\rotatebox{90}{XAI-based}}
& 0.005 & -0.26 & 0.11 & -0.00 & 0.38 & 23.99 & -0.21 \\
& 0.010 & -0.52 & 0.17 & -0.00 & 0.60 & 37.27 & -0.46 \\
& 0.020 & -0.78 & 0.21 & -0.01 & 0.75 & 46.65 & -0.78 \\
& 0.050 & -1.81 & 0.36 & -0.02 & 1.30 & 80.94 & -2.03 \\
\bottomrule
\end{tabular}
\vspace{-0.4cm}
\end{table}

\subsection{Numerical Results}

Tab.~\ref{tab:improvement_results} reports the percentage improvement over the baseline classifier for two deferring strategies: entropy-based rejection and the proposed XAI-based verifier, on dataset 1. Recall that class~1 corresponds to the ``good'' lightpath QoT; therefore, the most critical error is the false positive (i.e., approving a bad sample, $0 \rightarrow 1$).

Entropy rejection, which removes the most uncertain predictions according to predictive entropy, achieves substantial reductions in false positive rate (FPR), reaching up to $79.07\%$ at $q=0.02$. This confirms that predictive entropy effectively concentrates a large fraction of harmful errors within a small inspection budget. However, this reduction comes at a noticeable automation cost (e.g., $-2.06\%$ at $q=0.02$), with increasingly aggressive rejection as $q$ grows.

In contrast, the proposed XAI-based verifier achieves even stronger FPR reductions, up to $89.67\%$, while sacrificing significantly less automation (e.g., only $0.55\%$ at $q=0.02$). This indicates that explanation- and physics-aware verification provides more selective and targeted filtering than pure uncertainty rejection. Rather than deferring all uncertain samples, the verifier defers predictions whose explanations are incoherent, weakly anchored in operational features, or inconsistent with physical conditions (e.g., long paths or high span counts contributing positively to a ``good'' prediction).

Importantly, while entropy rejection slightly increases true positive rate (TPR), this effect largely stems from indiscriminate filtering of uncertain samples. In contrast, the proposed verifier preserves TPR almost unchanged while achieving stronger reductions in FPR. Such behavior is particularly desirable in safety-critical settings, where reducing dubious approvals must not compromise acceptance of genuinely good samples.

The results in Tab.~\ref{tab:improvement_results_exp2}, corresponding to dataset 2, further confirm the structured advantage of the XAI-based verifier. Entropy rejection again achieves strong FPR reductions (up to $78.92\%$ at $q=0.05$), but at substantially higher automation loss ($-5.06\%$). In contrast, the proposed verifier achieves comparable or superior FPR reductions with markedly smaller reductions in automation (e.g., $-1.81\%$ at $q=0.05$). This consistent pattern indicates that entropy operates as a broad uncertainty filter, whereas the proposed method performs structured risk discrimination. The distinction becomes clearer when examining TPR. Entropy rejection produces small positive shifts in TPR (up to $0.77\%$), reflecting non-specific filtering effects. Conversely, the coherence+physics verifier leaves TPR essentially unchanged (variations below $0.02\%$), indicating that it selectively targets risky approvals without degrading acceptance of truly good samples. Although minor increases in false negative rate (FNR) are observed under the proposed verifier, their magnitude remains negligible relative to the achieved FPR reductions. This confirms that the verifier primarily suppresses the critical error mode ($0 \rightarrow 1$) while preserving overall predictive stability.

These results suggest that integrating explanation coherence with domain-specific physical constraints may provide a structured mechanism to improve the risk--automation trade-off beyond uncertainty-based rejection alone. Unlike entropy, which treats uncertainty as a generic signal, the proposed verifier leverages domain-specific properties of the explanation, such as semantic consistency and physical plausibility, to differentiate between potentially risky approvals and merely uncertain predictions. The observed reductions in the critical $0 \rightarrow 1$ error mode, together with limited impact on automation levels, indicate that explanation-based signals carry actionable information that can be exploited for selective deferring. While the approach should be interpreted as a proof-of-concept rather than a finalized methodology, the findings highlight the potential of explanation-aware verification as a complementary safety layer alongside conventional uncertainty measures.

\section{Conclusion}
In this work, we introduced an explanation-based runtime verification framework that leverages model explanations by assessing their coherence together with physics-aware consistency checks. Experimental results demonstrate that the proposed approach enables selective rejection of high-risk approvals, substantially reducing the critical false positive rate while preserving a high automation level and maintaining stable true positive performance. The proposed methodology is complementary to uncertainty quantification and conformal prediction techniques, and can be naturally integrated with them to further strengthen decision reliability. More broadly, the findings point toward the relevance of structured, concept-aware learning paradigms such as concept bottleneck models for optical networks, where intermediate, physically meaningful representations could facilitate both interpretability and verifiable control.


\bibliographystyle{IEEEtran}
\bibliography{references}

\end{document}